\documentclass[11pt]{article}

\usepackage[preprint]{acl}

\usepackage{times}
\usepackage{latexsym}
\usepackage{amssymb}  
\usepackage[T1]{fontenc}

\usepackage[utf8]{inputenc}
\usepackage{graphicx}
\usepackage{microtype}

\usepackage{inconsolata}

\usepackage{graphicx}
\usepackage{multicol}
\usepackage{multirow}
\usepackage{xcolor}
\usepackage[most]{tcolorbox}
\usepackage{cuted}

\usepackage{booktabs}

\newtcolorbox{paperprompt}[2]{
  enhanced,
  arc=2mm,
  outer arc=2mm,
  boxrule=0.8pt,
  colframe=#1!75!black,
  colback=#1!8,
  coltitle=white,
  colbacktitle=#1!85!black,
  fonttitle=\bfseries,
  title={#2},
  titlerule=0pt,
  left=2mm,
  right=2mm,
  top=1.5mm,
  bottom=1.5mm,
}

\newenvironment{fullwidthprompt}[2]
{
  \begin{figure*}[t]
  \centering
  \begin{minipage}{0.95\textwidth}
  \begin{paperprompt}{#1}{#2}
}
{
  \end{paperprompt}
  \end{minipage}
  \end{figure*}
}

\title{DocTalkBN: A Novel Dataset of Expert Telemedicine Conversations in Bengali}

\author{
  \textbf{Anik Saha\textsuperscript{1,*}},
  \textbf{Fahmida Sultana Naznin\textsuperscript{1,*}},
  \textbf{Sadatul Islam Sadi\textsuperscript{1}},
\\
  \textbf{Ananya Shahrin Promi\textsuperscript{1}},
  \textbf{Wahid Al Azad Navid\textsuperscript{1}},
  \textbf{Rifat Shahriyar\textsuperscript{1}}
\\
\\
  \textsuperscript{1}Bangladesh University of Engineering and Technology (BUET)
\\
  \small{\textsuperscript{*}These authors contributed equally and are listed alphabetically.}
\\
  \small{
    \textbf{Correspondence:} \href{mailto:rifat@cse.buet.ac.bd}{rifat@cse.buet.ac.bd}
  }
}

\begin{document}
\maketitle

\begin{abstract}
Reliable medical conversational AI requires authentic expert--patient interaction data, yet such datasets remain scarce, especially for low-resource languages such as Bengali. We present DocTalkBN, a large-scale multimodal dataset of real-world expert telemedicine conversations in Bengali, collected from nationally broadcast telemedicine programs featuring board-certified physicians. DocTalkBN contains 557.63 hours of paired audio and text, 1,515 multi-turn patient calls, 10,274 host--doctor question--answer exchanges, totaling 1.7M tokens, spanning 26 medical specialties. Unlike prior resources derived from medical forums, written health content, or synthetic data, our dataset preserves the spontaneity, contextual richness, and spoken characteristics of authentic medical interactions in a low-resource setting. To support benchmark-driven research, we further construct three downstream tasks from the corpus, medical triage classification, advice safety evaluation, and medical named entity recognition, and benchmark a diverse set of large language models and encoder-based baselines. Our results show that DocTalkBN is a practically useful resource, particularly for clinically grounded reasoning tasks. We release this resource to facilitate future research on reliable medical NLP and safer, more culturally grounded healthcare systems for low-resource languages. Our source codes and dataset are publicly available at \url{https://anonymous.4open.science/r/doctalk}.
\end{abstract}

\section{Introduction}

\begin{figure}[t]
    \centering
    \includegraphics[width=0.4\textwidth]{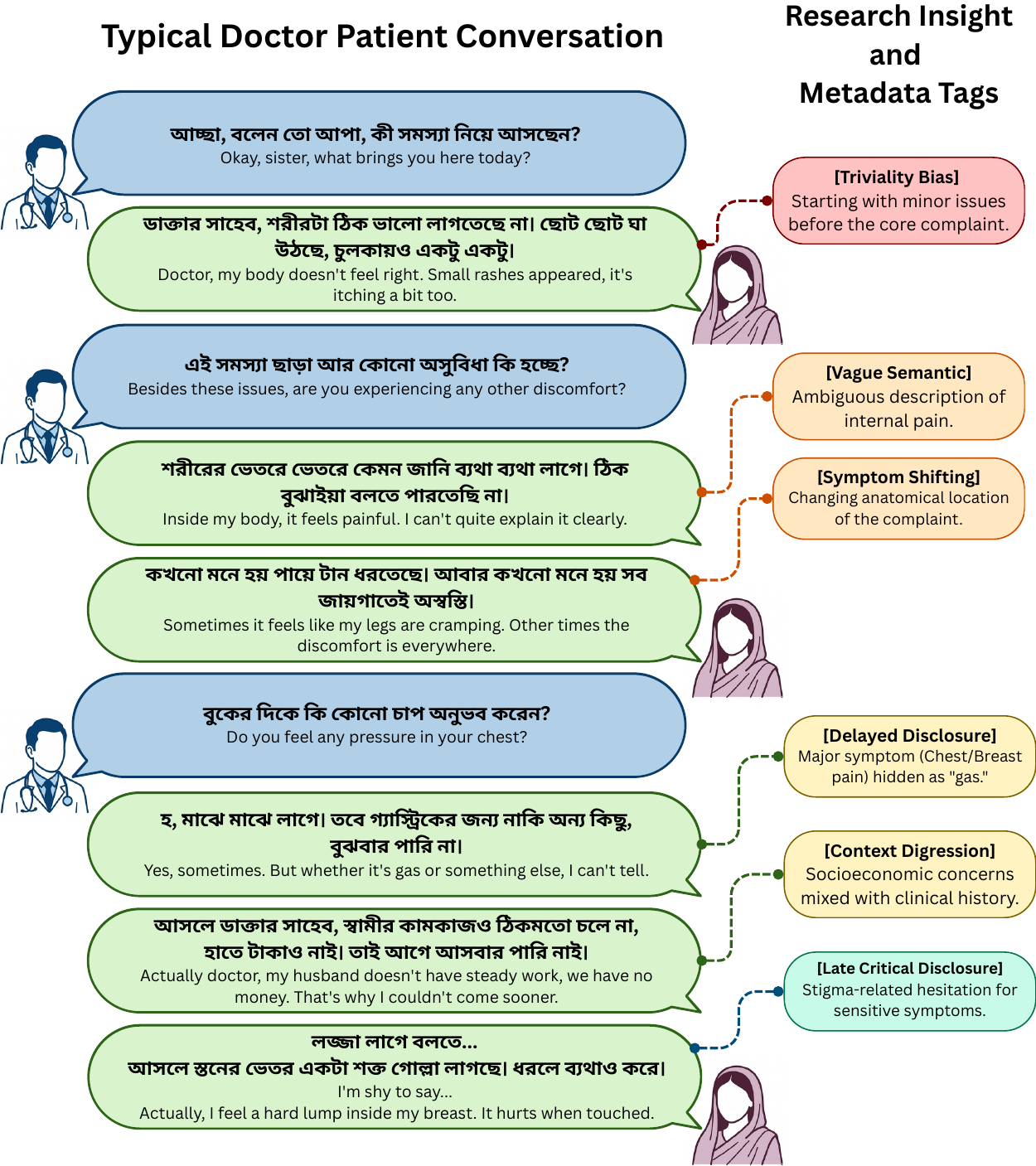}
    \caption{Dialogue between doctor and patient annotated with metadata tags.}
    \label{fig:doctor_patient_metadata}
\end{figure}

Conversational AI in healthcare has the potential to transform access and equity, providing expert guidance to millions with limited access to medical services \cite{uddin2025conversational}. Large language models (LLMs) along with rule-based and task-oriented agents, are rapidly reshaping healthcare communication across clinical and nonclinical settings \cite{jang-etal-2025-chatbot}. Automated systems often produce inaccurate, generalized, or biased responses due to reliance on user-generated prompts and limited publicly available training data, particularly in low-resource settings, which can reinforce existing health disparities and pose risks in safety-critical medical contexts \cite{maslenkova-etal-2025-building}. Building reliable healthcare AI depends on high-quality, authentic doctor–patient dialogue datasets; while resources like MedDialog provide large-scale conversations, they fail to capture the full complexity of real clinical interactions between doctor and patients \cite{zeng-etal-2020-meddialog}. This gap is critical in low-resource linguistic contexts, where cultural nuances, and spontaneous speech patterns complicate the training of culturally attuned, voice-enabled clinical agents for diverse populations (Figure~\ref{fig:doctor_patient_metadata}).
Authentic doctor-patient conversations are inherently messy, contextually rich, and diagnostically informative precisely because they deviate from fixed structure. Patients rarely present complaints in chronological order; instead, they interleave major symptoms with trivial concerns, omit critical temporal anchors, under-report negatives, repeat details, digress into family finances or emotional burdens, hesitate on sensitive topics, and deploy opaque regional expressions \cite{ben-abacha-etal-2023-overview}. Although these features pose challenges for automated analysis, they capture the very cues expert clinicians rely on to infer latent states and make safe, informed recommendations. Thus, conversational datasets that faithfully preserve both acoustic nuances and linguistic authenticity are not just valuable-they are essential for developing models capable of true clinical reasoning and empathetic understanding under uncertainty.

A rapidly growing body of work in medical NLP has yielded valuable resources, yet most remain confined to structured clinical articles/notes, curated forums, or de-identified EHR narratives-formats that inherently remove the spontaneity, prosodic cues, and expert grounding crucial for real-world clinical applications \citep{gao2022scoping, sazzed-2022-banglabiomed, ben-abacha-etal-2023-empirical, khan-etal-2023-nervous}. While recent collections of doctor–patient dialogues have begun to address this gap \citep{xu-etal-2022-realmeddial, saley-etal-2024-meditod}, they are generally limited in scale, restricted to text-only data, focused on single medical field, and mostly available in high-resource languages. To the best of our knowledge, there is no publicly available large-scale multimodal dataset of authentic medical conversations for low-resource settings. A high-quality dataset capturing real-world conversations, reflecting regional nuances, dialect variations, and typical communication challenges, would be extremely valuable. By incorporating multimodal data from interactions between experts and patients across different healthcare functions, this dataset can support applications such as medical response generation, clinical decision support, personalized patient care, and enhanced doctor–patient communication systems.

To address this gap, we introduce DocTalkBN, the first large-scale multimodal dataset of expert-grounded doctor–patient interactions in Bangla. Sourced from 1,934 videos of widely viewed, nationally broadcast telemedicine programs featuring board-certified specialists, DocTalkBN comprises $1,515$ multi-turn doctor–patient conversations and $10,274$ single-turn host–doctor question–answer exchanges, accompanied by 557.63 hours of time-aligned audio. These programs are not only popular-regularly attracting millions of viewers seeking trusted medical experts-but are institutionally reliable, as participating physicians are pre-vetted experts delivering unscripted, real-time advice. From this foundation we derive three clinically actionable downstream datasets-medical triage classification, advice safety evaluation, and medical named entity recognition-each constructed with rigorous LLM-assisted curation followed by multi-annotator human validation. The key contributions are summarized as follows:
\begin{itemize}
    \item We introduce DocTalkBN, the first large-scale multimodal dataset of expert-grounded Bangla doctor–patient conversations collected from nationally broadcast telemedicine programs, capturing authentic, unscripted clinical interactions with rich linguistic and acoustic characteristics.

    \item We develop a structured annotation and curation pipeline combining LLM-assisted processing with multi-annotator human validation to construct clinically meaningful benchmarks, enabling reliable evaluation of medical dialogue understanding in low-resource settings.

    \item We release three downstream benchmark tasks—medical triage classification, advice safety evaluation, and medical named entity recognition—demonstrating the potential of DocTalkBN for applications such as medical response generation, clinical decision support, and improved doctor–patient communication systems.
\end{itemize}

\begin{figure*}[htbp]
    \centering
    \includegraphics[width=0.9\textwidth]{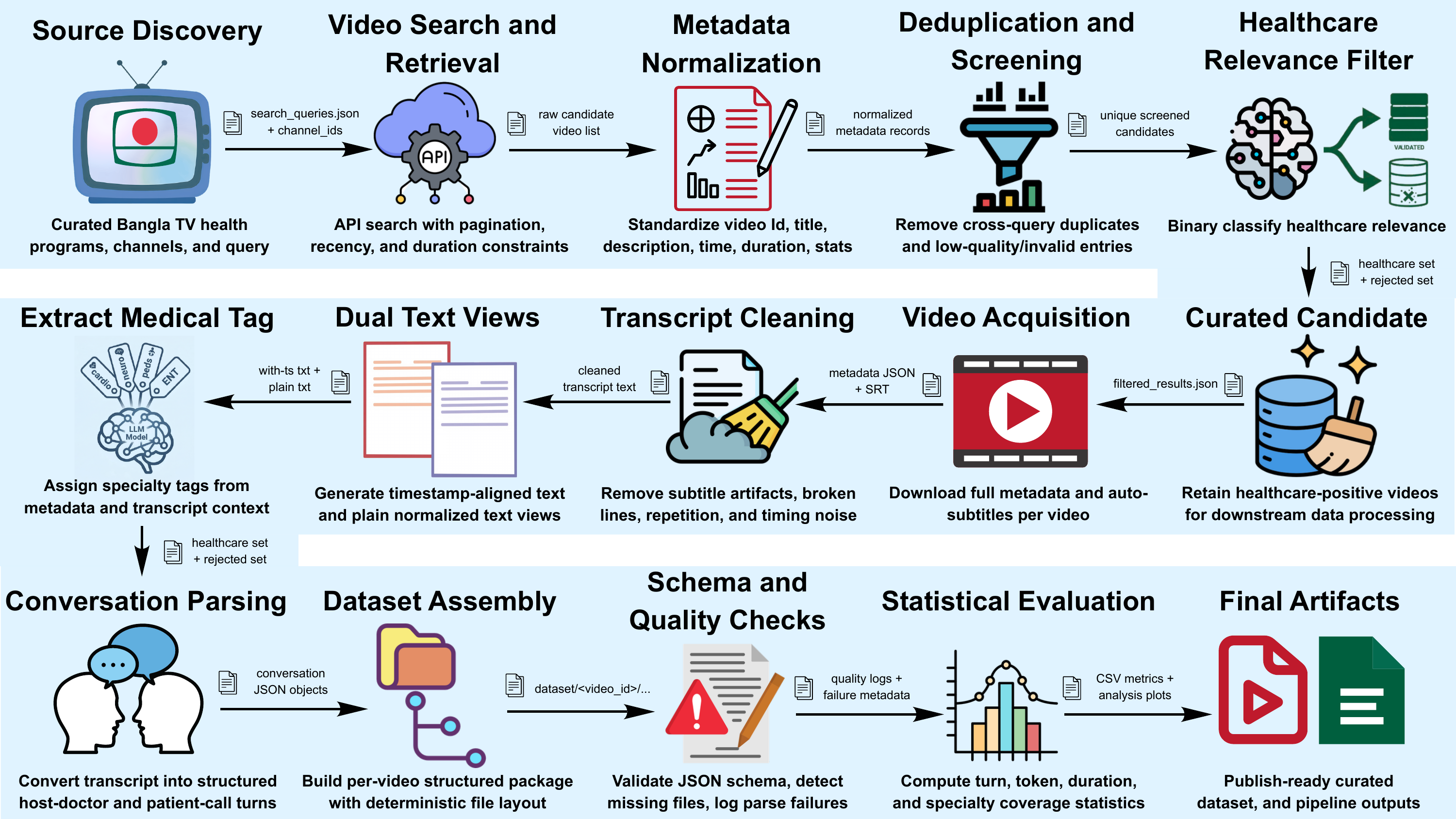}
    \caption{Overview for dataset curation and processing workflow.}
    \label{fig:data-curation-pipeline}
\end{figure*}

\section{Related Work}

Recent work in clinical natural language processing increasingly focuses on modeling patient–provider conversations, supported by datasets and downstreaming tasks for dialogues into structured medical documentation, such as MTS-Dialog~\cite{ben-abacha-etal-2023-empirical} and MEDIQA-Chat~\cite{benabacha2023mediqa}. Large-scale medical dialogue datasets include MedDialog~\cite{zeng-etal-2020-meddialog}, RealMedDial~\cite{xu-etal-2022-realmeddial}, and HealthCareMagic~\cite{pal2020meddialogue}, while task-oriented systems leverage PriMock57~\cite{liu2022primock}, ChiCCo~\cite{min2020summarizing}, CLINIC150~\cite{larson2019evaluation}, and MedDG~\cite{liu2022meddg}. Large language models generate or augment dialogues (e.g., NoteChat~\cite{wang2023notechat}) or expand limited summarization training data~\cite{schlegel2023pulsar}. Recent work explores discourse and reasoning, physician intent aligned with SOAP frameworks~\cite{roehr2025intent}, LLM-based clinical note generation~\cite{sharma-etal-2023-team,giorgi-etal-2023-wanglab}, and robustness on out-of-domain SOAP notes~\cite{chen-hirschberg-2024-exploring}.

For low-resource languages like Bangla, biomedical NER datasets such as BanglaBioMed~\cite{sazzed-2022-banglabiomed} and Bangla‑HealthNER~\cite{khan-etal-2023-nervous} exist, alongside conversational datasets like BanglaCHQ‑Summ for abstractive summarization of patient–doctor interactions~\cite{banglachq_summ}. Additional Bangla resources include healthcare chat corpora, domain-specific health paraphrases~\cite{banglahealth_paraphrase}, entity recognition datasets like Bangla‑MedER~\cite{bangla_meder}, clinical symptom reports, and multimodal resources such as MedBanglaTrust3~\cite{medbanglatrust3}. However, these are mostly text-based or scripted and lack authentic, spontaneous patient–doctor dialogues, highlighting the need for natural Bangla medical conversation datasets.

\section{Dataset Statistics}

Table~\ref{tab:dataset_comparison} compares existing medical dialogue datasets with DocTalkBN. Most prior datasets are derived from online text sources. Only RealMedDial contains real doctor--patient consultations, but it is limited to Chinese. For Bangla, datasets such as BanglaBioMed and Bangla-HealthNER~\cite{sazzed-2022-banglabiomed} focus on named entity recognition, while BanglaCHQ-Summ~\cite{banglachq_summ} targets medical query summarization. These datasets are collected from medical forums and are purely textual. In contrast, DocTalkBN contains both audio and textual data from live Bangla doctor--patient consultations, capturing the natural complexity of real clinical interactions in a low-resource setting.








\begin{table*}[t]
\centering
\caption{Comparison between our dataset and existing medical dialogue datasets.}
\label{tab:dataset_comparison}
\resizebox{0.9\textwidth}{!}{%
\begin{tabular}{lllrrrl}
\hline
Dataset Name & Source Type & Language & \#Dialogues & \#Utterances & \#Diseases & Department \\
\hline
MZ~\cite{wei-etal-2018-task} & Online Text & Chinese & 710 & - & 4 & Pediatrics \\
DX~\cite{xu2019end} & Online Text & Chinese & 527 & 2,186 & 5 & Pediatrics \\
CMDD~\cite{lin2019enhancing} & Medical Forum & Chinese & 2,067 & 87,005 & 4 & Pediatrics \\
MedDG~\cite{liu2022meddg} & Online Text & Chinese & 17,864 & 385,951 & 12 & Gastroenterology \\
MedDialog-CN~\cite{pal2020meddialogue} & Online Text & Chinese & 3,407,494 & 11,260,564 & 172 & 51 Departments \\
RealMedDial~\cite{xu-etal-2022-realmeddial} & Video Clips & Chinese & 2,637 & 24,255 & 55 & 17 Departments \\
BanglaBioMed~\cite{sazzed-2022-banglabiomed} & Health Article & Bangla & -- & 818 & -- & -- \\
Bangla-HealthNER~\cite{sazzed-2022-banglabiomed} & Medical Forum & Bangla & -- & 144,136 & -- & -- \\
BanglaCHQ-Summ~\cite{banglachq_summ} & Medical Forum & Bangla & 2,350 & -- & 32 & -- \\
\hline
DocTalkBN (Ours) & Live Consultation & Bangla & 11,789 & -- & -- & 26 Departments \\
\hline
\end{tabular}%
}
\end{table*}

Table \ref{tab:detailed_dataset_statistics} summarizes further key statistics including dataset characteristics, patient-doctor calls, and host-doctor conversations.
\begin{table}[h]
\centering
\caption{Detailed Dataset Statistics}
\label{tab:detailed_dataset_statistics}
\resizebox{\columnwidth}{!}{%
\begin{tabular}{l|c}
\hline
\multicolumn{2}{c}{\textbf{Overall Dataset Statistics}} \\ \hline
Dataset Duration (Hours) & 557.63 \\
Unique Medical Specialties & 26 \\
Patient Calls Proportion (\%) & 12.9 \\
Host-Doctor QA Proportion (\%) & 87.1 \\ \hline

\multicolumn{2}{c}{\textbf{Patient-Doctor Calls}} \\ \hline
Total Patient-Doctor Calls & 1,515 \\
Total Tokens & 244,027 \\
Total Turns & 4,857 \\
Avg Turns per Call & 3.2 \\
Max Turns in a Call & 14 \\
Avg Tokens per Turn & 50.2 \\
Total Doctor Tokens & 156,573 (64.2\%) \\
Total Patient Tokens & 87,454 (35.8\%) \\
Avg Doctor Tokens / Call & 103.3 \\
Avg Patient Tokens / Call & 57.7 \\ \hline

\multicolumn{2}{c}{\textbf{Host-Doctor QA}} \\ \hline
Total Host-Doctor QA Sessions & 10,274 \\
Total Tokens & 1,427,183 \\
Total Turns & 20,553 \\
Avg Turns per QA Session & 2.0 \\
Max Turns in a QA Session & 4 \\
Avg Tokens per QA Session & 138.9 \\
Avg Tokens per Turn & 69.4 \\ \hline
\end{tabular}%
}
\end{table}

Table~\ref{tab:downstream} presents statistics regarding the downstream evaluation datasets.

\begin{table}[htbp]
\centering
\small
\caption{Downstream Benchmark Datasets \& Class Distributions}
\label{tab:downstream}
\resizebox{\columnwidth}{!}{%
\begin{tabular}{l r r r l r}
\hline
\textbf{Downstream Task} & \textbf{\#Videos} & \textbf{Samples} & \textbf{Total Labels} & \textbf{Class} & \textbf{Count} \\
\hline


\multirow{2}{*}{Advice Safety} 
& \multirow{2}{*}{1{,}744} & \multirow{2}{*}{8{,}689} & \multirow{2}{*}{29{,}957}
& SAFE & 22{,}513 \\
& & & & HARM & 7{,}444 \\
\hline

\multirow{7}{*}{Medical NER} 
& \multirow{7}{*}{1{,}842} & \multirow{7}{*}{11{,}239} & \multirow{7}{*}{105{,}697}
& DIS & 31{,}680 \\
& & & & SYM & 23{,}385 \\
& & & & ANA & 21{,}624 \\
& & & & TRT & 13{,}802 \\
& & & & DRG & 7{,}608 \\
& & & & TST & 6{,}647 \\
& & & & MED & 951 \\
\hline

\multirow{4}{*}{Triage} 
& \multirow{4}{*}{490} & \multirow{4}{*}{1{,}515} & \multirow{4}{*}{1{,}515}
& REF & 1{,}028 \\
& & & & SELF & 258 \\
& & & & ROUT & 185 \\
& & & & URG & 44 \\
\hline

\end{tabular}%
}
\end{table}

Table~\ref{tab:dataset_stats} provides a detailed breakdown of the dataset statistics. The dataset contains a total of 1,934 videos with a cumulative duration of 557.63 hours. It includes 10,274 host doctor question-answer exchanges and 1,515 patient calls. 
Each dialogue spans 2 to 14 turns, covering interactions from 26 distinct medical departments. 

\begin{table}[ht]
\centering
\caption{Statistics of the Bangla DocTalkBN dataset}
\label{tab:dataset_stats}
\resizebox{\columnwidth}{!}{%
\begin{tabular}{l|r}
\hline
\# Dialogues & 11,789 \\
\# Utterances &  25,410
\\
\# Tokens & 1,671,210 \\
\hline
Avg. \# Utterances in a Dialogue & 2.16 \\
Max \# Utterances in a Dialogue & 14 \\
Min \# Utterances in a Dialogue & 1 \\
\hline
Avg. \# Tokens in an Utterance & 65.8  \\
Max \# Tokens in an Utterance & 418 \\
Min \# Tokens in an Utterance & 2 \\
\hline
\end{tabular}%
}
\end{table}

\section{Methodology}

\subsection{Data Collection and Preprocessing}

We construct DocTalkBN from publicly available telemedicine programs broadcast on national television channels in Bangladesh and later uploaded to YouTube. These programs provide a valuable source of expert-grounded medical interactions, including both doctor--patient consultations and host--doctor question--answer exchanges. To build the corpus, we first search the relevant YouTube channels over a five-year period and retrieve candidate videos. We then use Qwen3-30B-Instruct ($L_1$) \cite{Arxiv-Qwen3-tech-report} to filter videos based on their titles and descriptions, retaining only those that are relevant to the medical domain. The same model is further used to assign medical specialty tags, such as cardiology, dermatology, and gastroenterology, which helps organize the collection across a diverse set of clinical topics.

For each selected video, we obtain the automatically generated Bangla subtitles provided by YouTube. While these transcripts are noisy and reflect spontaneous spoken language, they preserve the essential medical content and conversational flow. We next apply a sequence of preprocessing steps, including segmenting long transcripts, removing non-conversational content such as introductions and advertisements, and normalizing transcription artifacts. Five independent human annotators reviewed the audio and subtitles to validate the transcriptions. Annotators were instructed to verify conversational fidelity, medical content preservation, speaker continuity, and major transcription errors. If at least two annotators rated a transcription as poor in quality, it was revised by a separate independent annotator. A total of 36 total transcriptions required such human revision. Next, we remove personally identifiable information (e.g., names and locations) using the BNLP Toolkit \footnote{\url{https://pypi.org/project/bnlp-toolkit/}}. Finally, we use Gemini-3-Flash ($L_2$) to reconstruct structured conversations by identifying speakers with timestamps, separating turns, and classifying interactions as either doctor--patient dialogues or host--doctor question--answer exchanges. This process yields DocTalkBN, a dataset of paired audio clips and expert-grounded medical conversations for downstream research.

\subsection{Data Curation for Downstream Tasks}

We next derive three downstream task datasets from DocTalkBN using Gemini-2.5-Flash ($L_3$), followed by human validation. We first iteratively refine the prompt over five rounds, with three independent human annotators reviewing the outputs of $L_3$ at each stage. After finalizing the prompt, we generate the downstream datasets using this updated version. The finalized dataset generation prompts are provided in Appendix. Each resulting sample is then reviewed by three independent annotators. Samples rated as poor quality by at least two of the three annotators are subsequently revised by a separate annotator. In total, 17 samples required human corrections ranging from NER span edits to classification corrections. For the discussion below, let $C_p = (P_1, D_1, P_2, \ldots, P_n, D_n)$ denote a multi-turn doctor--patient conversation, and let $C_h = (H, D)$ denote a host--doctor exchange. We define $PatientProfile(C_p)$ as the patient state inferred from all turns preceding the final doctor response $D_n$. Thus, the patient profile is constructed from the patient’s initial complaint and subsequent replies to the doctor’s follow-up questions, while excluding the final recommendation itself. This formulation allows us to build evaluation settings that focus on what can be inferred from the presented symptoms and history, rather than from the doctor’s concluding advice.

\paragraph{Medical Triage.} Medical triage aims to determine the appropriate level of care for a patient based on the severity and urgency of the condition. For each doctor--patient conversation $C_p$, we prompt $L_3$ to map the doctor’s final recommendation $D_n$ into one of four triage categories: \texttt{REASSURANCE\_SELF\_CARE} (SELF), \texttt{ROUTINE\_OUTPATIENT\_VISIT} (ROUT), \texttt{INVESTIGATION\_OR\_SPECIALIST\_REFERRAL} (REF), and \texttt{URGENT\_EMERGENCY\_CARE} (URG). This categorization is inspired by \cite{Gatto-2026-Medical-Triage}. This process yields an annotated dataset of the form $\{(P, T)\}$, where $P = PatientProfile(C_p)$ and $T$ is the corresponding triage label. Notably, although the full conversation $C_p$ is used during dataset construction to recover the doctor’s intended recommendation, inference-time evaluation uses only the patient profile $P$ as input. This setup makes the task clinically meaningful, as the model must infer the appropriate treatment pathway from the patient’s history alone.

\paragraph{Advice Safety Evaluation.}
This task focuses on distinguishing safe from harmful courses of action for a given medical condition. As in the triage setting, we first derive a patient profile $P$ from each doctor--patient conversation $C_p$. We then prompt $L_3$ to decompose the doctor’s final recommendation $D_n$ into a set of modular advice units, which are grouped into $A_p$ and $A_n$, denoting safe and harmful actions, respectively. For example, for a patient with fever and dehydration, ``drink adequate fluids'' would be labeled as a safe action, whereas ``ignore persistent breathing difficulty and remain at home'' would be labeled as harmful. Where applicable, we also extract such condition--advice pairs from host--doctor exchanges $C_h$. Based on this annotation scheme, we formulate the task as follows: given a patient profile $P$ and an action $a \in A_p \cup A_n$, the model must determine whether the action is safe or harmful. This setup enables a fine-grained evaluation of medical advice safety at the level of individual recommendations rather than entire responses.
 
\paragraph{Medical Entity Recognition.}
We further formulate a medical named entity recognition task in the context of Bangla conversational healthcare data. Inspired by \cite{khan-etal-2023-nervous, sazzed-2022-banglabiomed}, we define seven entity categories: \texttt{SYMPTOM\_SIGN} (SYM), \texttt{DISEASE\_CONDITION} (DIS), \texttt{DRUG\_MEDICATION} (DRG), \texttt{TEST\_INVESTIGATION} (TST), \texttt{TREATMENT\_PROCEDURE} (TRT), \texttt{ANATOMY\_BODY\_PART} (ANA), and \texttt{MEDICAL\_SPECIALTY} (MED). We then use $L_3$ to identify and extract mentions of these entity types from the curated conversations, followed by human validation. This task supports structured understanding of naturally occurring Bangla medical dialogue and provides a useful benchmark for information extraction in low-resource clinical language settings. Figure \ref{fig:data-curation-pipeline} presents an illustration of the data curation workflow.

\section{Experiments on Downstream Tasks}

In this section, we report the experimental setup and results on the downstream tasks.

\subsection{Experimental Setup}

All experiments are conducted on a machine equipped with a single NVIDIA RTX 4090 GPU (24GB VRAM). When feasible, we run open LLMs locally using Ollama.\footnote{\url{https://ollama.com/}} In other cases, for both open and closed models, we use the OpenRouter API\footnote{\url{https://openrouter.ai/docs/api/reference/overview}} for inference. We use an 80:10:10 split for training, validation, and test, respectively. During LLM inference, we use a consistent decoding setup with temperature $=0.7$, top\_p $=0.9$, top\_k $=35$, and repeat\_penalty $=1.1$ whenever supported. For few-shot prompting, we typically use 3 to 5 examples, depending on the task. For BERT-based experiments, we fine-tune the models for 100 epochs with a batch size = $16$, warm-up steps = $200$, and early stopping (patience = $10$). 

\subsection{Baselines}
We benchmark a diverse set of LLMs, including GPT-4o, GPT-5 mini, Llama-3.3-70B-Instruct \cite{llama3modelcard}, Qwen3.5-Flash \cite{qwen3.5}, DeepSeek-V3.2 \cite{deepseekai2025deepseekv32}, and Gemma-3-27B-IT \cite{gemma_2025}. These baselines span both proprietary and open-weight models, and cover general-purpose as well as reasoning-oriented model families. For the medical named entity recognition task, we additionally fine-tune two encoder-based baselines, namely BanglaBERT \cite{bhattacharjee-etal-2022-banglabert} and mmBERT \cite{marone2025mmbert}.

\subsection{Evaluation Metrics}

We now present our evaluation metrics for each downstream task.

\paragraph{Medical Triage.}
We formulate triage as a four-way single-label classification problem and report the macro-F1 score. Let $TP_c$, $FP_c$, and $FN_c$ denote the true positives, false positives, and false negatives for class $c$. Then
\[
\mathrm{Precision}_c = \frac{TP_c}{TP_c + FP_c}
\]
\[
\mathrm{Recall}_c = \frac{TP_c}{TP_c + FN_c},
\]
\[
F1_c = \frac{2 \cdot \mathrm{Precision}_c \cdot \mathrm{Recall}_c}{\mathrm{Precision}_c + \mathrm{Recall}_c}.
\]
Macro-F1 is computed as the unweighted mean of class-wise F1 values. Since the triage dataset is imbalanced, we use macro-F1 as the primary metric.

\paragraph{Advice Safety Evaluation.}
Advice safety is formulated as a binary classification task over individual recommendation units, with labels \texttt{SAFE} and \texttt{HARMFUL}. Since it is also a classification problem, we use the same metric Macro F1 as above.

\paragraph{Medical Named Entity Recognition.}
For medical NER, we evaluate entity extraction using the \textit{strict} matching scheme over the seven entity categories in our benchmark. Under strict evaluation, a predicted entity is counted as correct only if both its span boundary and entity label exactly match the gold annotation. Let $COR$, $ACT$, and $POS$ denote the numbers of correctly predicted entities, predicted entities, and gold entities, respectively. Precision and recall are computed as
\[
\mathrm{Precision} = \frac{COR}{ACT}, \qquad
\mathrm{Recall} = \frac{COR}{POS},
\]
and strict F1 is defined as the harmonic mean of precision and recall. We report the macro-F1 under this criterion.

\subsection{Results}

We now present the results on the downstream tasks across different models and evaluation settings. The prompts used for LLM inference on each task are provided in Appendix.

\subsubsection{Results on Medical Triage}

\begin{table}[t]
\centering
\caption{Triage classification performance across models and settings. Best values are shown in \textbf{bold} and second best values are \underline{underlined}.}
\label{tab:triage-aggregate-results}
\resizebox{\columnwidth}{!}{%
\begin{tabular}{llr}
\hline
Model & Setting & Macro F1 \\
\hline
DeepSeek V3.2 & Zero-shot & 0.320 \\
DeepSeek V3.2 & Few-shot & 0.383 \\
Gemma 3 27B Instruct & Zero-shot & 0.412 \\
Gemma 3 27B Instruct & Few-shot & 0.422 \\
GPT-4o & Zero-shot & 0.425 \\
GPT-4o & Few-shot & \underline{0.448} \\
GPT-5 Mini & Zero-shot & 0.414 \\
GPT-5 Mini & Few-shot & 0.427 \\
Llama 3 70B Instruct & Zero-shot & \textbf{0.463} \\
Llama 3 70B Instruct & Few-shot & 0.443 \\
Qwen 3.5 Flash & Zero-shot & 0.363 \\
Qwen 3.5 Flash & Few-shot & 0.416 \\
\hline
\end{tabular}
}
\end{table}

We report macro F1 scores for triage classification under zero-shot and few-shot prompting in Table~\ref{tab:triage-aggregate-results}. Despite involving only four target categories, the task remains challenging for all models, with the best performance reaching an F1 score of 0.463, achieved by Llama~3~70B Instruct in the zero-shot setting. GPT-4o delivers the second-best result, obtaining an F1 score of 0.448 in the few-shot setting. In contrast, DeepSeek~V3.2 and Qwen~3.5 Flash perform less competitively overall, with their best few-shot scores reaching 0.383 and 0.416, respectively. We also observe that adding in-context examples generally improves performance across models, with Llama~3~70B Instruct being the only exception, where the few-shot setting leads to a slight decline relative to zero-shot prompting.

\subsubsection{Results on Advice Safety Evaluation}

\begin{table}[t]
\centering
\caption{Advice safety classification performance across models and settings. Best values are shown in \textbf{bold} and second best values are \underline{underlined}.}
\label{tab:advice-safety-aggregate-results}
\resizebox{\columnwidth}{!}{%
\begin{tabular}{llr}
\hline
Model & Setting & Macro F1 \\
\hline
DeepSeek V3.2 & Zero-shot & 0.937 \\
DeepSeek V3.2 & Few-shot & 0.949 \\
Gemma 3 27B Instruct & Zero-shot & 0.951 \\
Gemma 3 27B Instruct & Few-shot & 0.955 \\
GPT-4o & Zero-shot & \underline{0.969} \\
GPT-4o & Few-shot & \textbf{0.976} \\
GPT-5 Mini & Zero-shot & 0.929 \\
GPT-5 Mini & Few-shot & 0.931 \\
Llama 3 70B Instruct & Zero-shot & 0.930 \\
Llama 3 70B Instruct & Few-shot & 0.924 \\
Qwen 3.5 Flash & Zero-shot & 0.948 \\
Qwen 3.5 Flash & Few-shot & 0.951 \\
\hline
\end{tabular}
}
\end{table}

Table~\ref{tab:advice-safety-aggregate-results} presents the macro F1 scores for advice safety classification under zero-shot and few-shot prompting. In contrast to the medical triage task, all models achieve consistently strong performance on this benchmark, with macro F1 values exceeding 0.92 across all settings. GPT-4o performs best overall, reaching 0.976 in the few-shot setting and 0.969 in the zero-shot setting, indicating both strong baseline capability and a clear benefit from in-context examples. Gemma~3~27B Instruct and Qwen~3.5 Flash also deliver highly competitive results, each attaining 0.955 and 0.951, respectively, under few-shot prompting. DeepSeek~V3.2 likewise benefits from few-shot examples, improving from 0.937 to 0.949. More broadly, few-shot prompting yields modest but consistent gains for most models, suggesting that the advice safety task is comparatively well aligned with current LLM capabilities and can be further strengthened through limited task-specific exemplars. A notable exception is Llama~3~70B Instruct, whose performance decreases slightly from 0.930 in the zero-shot setting to 0.924 in the few-shot setting as was the case with medical triage as well. Overall, these results indicate that advice safety evaluation is substantially easier for current models than medical triage classification, while still revealing meaningful differences in robustness across model families.

\subsubsection{Results on Medical Entity Recognition}

\begin{table}[t]
\centering
\caption{Performance on Medical NER across models and prompting/training settings. Best values are shown in \textbf{bold} and second best values are \underline{underlined}.}
\label{tab:ner-aggregate-results}
\resizebox{\columnwidth}{!}{%
\begin{tabular}{llr}
\hline
Model & Setting & Strict F1 (Macro Avg.) \\
\hline
BanglaBERT & Fine-tuned & \underline{0.742} \\
mmBERT & Fine-tuned & 0.707 \\
DeepSeek V3.2 & Zero-shot & 0.694 \\
Gemma 3 27B Instruct & Zero-shot & 0.585 \\
GPT-4o & Zero-shot & 0.581 \\
GPT-4o & Few-shot & 0.575 \\
GPT-5 Mini & Zero-shot & 0.704 \\
GPT-5 Mini & Few-shot & 0.703 \\
Llama 3 70B Instruct & Zero-shot & 0.489 \\
Llama 3 70B Instruct & Few-shot & 0.477 \\
Qwen 3.5 Flash & Zero-shot & \textbf{0.743} \\
Qwen 3.5 Flash & Few-shot & 0.741 \\
\hline
\end{tabular}
}
\end{table}

Table~\ref{tab:ner-aggregate-results} reports aggregate strict macro F1 scores for medical NER across both fine-tuned encoder models and prompted LLMs. Overall, the results show that the task remains challenging, but several models achieve competitive performance. Qwen~3.5 Flash attains the best overall score with a strict F1 of 0.743 in the zero-shot setting, narrowly outperforming the fine-tuned BanglaBERT baseline, which achieves 0.742. Among the supervised encoder-based models, BanglaBERT substantially surpasses mmBERT, whose fine-tuned performance reaches 0.707, highlighting the advantage of stronger Bengali-specific representations for this task. Among the LLMs, DeepSeek~V3.2 and GPT-5 Mini perform comparatively well, obtaining 0.694 and 0.704 in the zero-shot setting, respectively, whereas Gemma~3~27B Instruct, GPT-4o, and Llama~3~70B Instruct trail behind by a wider margin. Notably, few-shot prompting does not improve performance for the models evaluated in both settings; instead, it yields either marginal decreases or near-identical results. This pattern suggests that, for medical NER, careful span extraction remains difficult to improve through a small number of in-context examples, while strong fine-tuned encoders and some zero-shot LLMs already provide a more effective inference.

\section{Findings and Error Analysis}

\paragraph{Medical triage remains the most challenging task.}
Medical triage proves to be the hardest of the three tasks, with only moderate performance even from the strongest models. This likely reflects the need for broader clinical reasoning over incomplete and conversational patient histories.

\paragraph{Frequent confusion between routine visit and referral decisions.}
Models often confuse \texttt{ROUTINE\_OUTPATIENT\_VISIT} with \texttt{INVESTIGATION\_OR\_SPECIALIST\_REFERRAL}. The distinction is subtle and usually depends on the implied level of escalation in the doctor's recommendation.

\paragraph{Advice safety is comparatively easier for current models.}
Advice safety is much easier for current models than triage. This task often depends on direct local cues, making the decision more straightforward.

\paragraph{NER errors often involve over-predicting generic medical words.}
In NER, models often label broad medical words such as disease, treatment, or medicine as entities. This suggests a reliance on surface-level medical association rather than the annotation rules.

\paragraph{Boundary errors remain a major challenge in NER.}
Models also frequently miss the full span of an entity and extract only part of it.




\section{Conclusion}

We introduced DocTalkBN, a large-scale multimodal dataset of real-world expert telemedicine conversations in Bengali, designed to support research on medical dialogue understanding in a low-resource language setting. Collected from nationally broadcast telemedicine programs, the dataset captures authentic doctor--patient interactions and host--doctor question--answer exchanges, preserving the linguistic variability, conversational structure, and spoken characteristics of naturally occurring clinical communication. We further derived three benchmark tasks from the corpus, medical triage classification, advice safety evaluation, and medical named entity recognition, and evaluated a diverse set of large language models and encoder-based baselines on these tasks. Our results show that while current models perform strongly on some aspects of medical dialogue understanding, clinically grounded reasoning tasks such as triage remain challenging. We hope DocTalkBN will serve as a valuable resource for advancing reliable medical NLP, conversational AI, and safer healthcare technologies for Bengali and other low-resource languages.

\section{Limitations and Future Work}

While our current study focuses on text-based benchmarking under zero-shot prompting, few-shot prompting and encoder fine-tuning, DocTalkBN also enables several broader research directions. In particular, the dataset includes paired audio along with text, which can support future work on medically grounded Bangla ASR in low-resource settings. Beyond advice safety classification, it can also facilitate evaluation of advice generation, such as producing safe and appropriate recommendation units directly from a patient profile. Moreover, the benchmark can be extended to study broader medical reasoning abilities of models in low-resource settings, including retrieval-augmented generation, web-assisted reasoning, tool use, and more general agentic capabilities.

\section{Ethical Considerations}

This study was conducted in accordance with institutional ethical guidelines. The collection of data from publicly accessible Youtube videos was approved by the Institutional Ethics Review Board. All identifying information of the patients and doctors was anonymized.

\section*{Acknowledgments}

We used large language models only for text formatting and presentation-related assistance. All data collection, annotation design, verification, analysis, and scientific conclusions were carried out by the authors.

\bibliography{references}

\appendix

\section{Appendix}
\label{sec:appendix}

\subsection{Hyperparameter details for finetuning}

Table~\ref{tab:ner-hyperparameters} presents the hyperparameter details of finetuning.

\begin{table}[t]
\centering
\small
\begin{tabular}{lcc}
\toprule
\textbf{Hyperparameter} & \textbf{BanglaBERT} & \textbf{mMBERT} \\
\midrule
Learning rate         & $3\times10^{-5}$ & $2\times10^{-5}$ \\
Epochs                & 100               & 100 \\
Train batch size      & 8                 & 4 \\
Evaluation batch size & 16                & 8 \\
Gradient accumulation & 2                 & 4 \\
Warmup steps          & 200               & 200 \\
Early stopping        & 10                & 2 \\
Weight decay          & 0.01              & 0.01 \\
FP16                  & No                & Yes \\
Logging steps         & 50                & 50 \\
\bottomrule
\end{tabular}
\caption{Hyperparameter settings used for fine-tuning the BanglaBERT and mMBERT models.}
\label{tab:ner-hyperparameters}
\end{table}

\subsection{Prompts}
In this section, we present the prompts used in different stages of data curation and evaluation. Where appropriate, we abridge the prompts and replace examples with placeholders. The full versions are available in our public code repository.

\begin{fullwidthprompt}{blue}{Healthcare Content Classification Prompt}
\textbf{[SYSTEM]} You are a reliable and deterministic classification
system. Follow the instructions exactly as provided. Do not add
explanations, reasoning, or extra text. Return only a valid JSON object
in the specified format.

\medskip

\textbf{[ROLE]} You are a strict healthcare content classifier.

\textbf{[TASK]} Determine whether a YouTube video's title and description
indicate healthcare-related content.

Healthcare-related content includes medicine or medical advice, physical
or mental health topics, diseases, symptoms, diagnosis, treatment, health
tips, wellness, nutrition, and fitness.

Non-healthcare content includes entertainment, gaming, vlogs, lifestyle,
political news, technology, travel, and general news unrelated to health.

A useful prior is that some targeted television program and channel names
are strongly associated with healthcare content:

\textit{<healthcare-oriented program/channel names>.}

However, these same channels may also publish non-healthcare videos.

If the title and description conflict, prefer the description when it
clearly indicates a healthcare program. For example:

\textit{<1 example where a news-like title appears with a healthcare-program
description and should be labeled healthcare>.}

\textbf{[OUTPUT FORMAT]} Return only:
\texttt{\{"healthcare": true | false\}}

\textbf{[EXAMPLES]} \textit{<6 title/description classification examples, including 3 positive
and 3 negative cases>.}
\end{fullwidthprompt}

\begin{fullwidthprompt}{red}{Medical Tag Extraction Prompt}
\textbf{[SYSTEM]} You are a reliable and deterministic healthcare video
metadata extraction system. Your current task is medical tag extraction.
Follow the instructions exactly. Do not add explanations, reasoning, or
extra text. Return only a valid JSON object in the specified format.

\textbf{[ROLE]} You are a healthcare video metadata extractor.

\textbf{[TASK]} Analyze the title and description of a healthcare video and
assign the most relevant medical specialty tags.

\textbf{[ALLOWED TAGS]} Choose only from a fixed tag set covering major
medical specialties and health domains:
\textit{<Allowed tags with English/Bangla descriptions, including
cardiology, neurology, gastroenterology, pulmonology, endocrinology,
nephrology, hepatology, dermatology, gynecology, obstetrics, pediatrics,
psychiatry, orthopedics, rheumatology, ophthalmology, ENT, urology,
infectious-disease, nutrition, oncology, general-medicine,
preventive-care, sexual-health, emergency-care, alternative-medicine,
and dentistry>.}

\textbf{[RULES]} Assign 1--2 tags describing the primary medical topics
(maximum 3 if multiple specialties are clearly involved). Use only tags
from the allowed list. Prefer more specific specialties over generic ones.
Use \texttt{general-medicine} only when no specific specialty fits, and use
\texttt{preventive-care} only when no more specific specialty is suitable.

\textbf{[OUTPUT FORMAT]} Return only:
\texttt{\{"tags": ["tag1", "tag2"]\}}

\textbf{[EXAMPLES]}
\textit{<5 title/description tagging examples covering single-specialty,
multi-specialty, and preventive-care/emergency-care cases>.}
\end{fullwidthprompt}

\begin{fullwidthprompt}{green}{Medical Conversation Parsing Prompt}
\textbf{[SYSTEM]} You are a data annotation assistant specialized in
medical dialogue datasets. Follow the instructions exactly. Return only
valid JSON in the specified format. Do not add explanations, comments, or
extra text. If the transcript is ambiguous, make the safest reasonable
interpretation without inventing medical information.

\textbf{[TASK]} Given a raw Bangla transcript from Bangladeshi public TV
health programs, parse it into a clean structured medical conversation
dataset suitable for dialogue analysis.

\textbf{[INPUT CHARACTERISTICS]} The transcript may contain host--doctor
discussions, real patient call-ins, doctor follow-up questions, subtitle
timestamps, repetitions, broken lines, and transcription artifacts.
Preserve medical meaning only; do not preserve TV-show framing, greetings,
or introductions.

\textbf{[CONTEXT]} In these programs, a host asks general health questions
to a doctor, patients may call in with symptoms, doctors may ask
supplementary clarification questions, and the host may paraphrase patient
statements. The symbol \texttt{>>} often indicates a speaker shift, but is
not fully reliable; infer structure using linguistic and medical context.

\textbf{[OUTPUT FORMAT]} Return only a valid JSON array of conversation
objects with fields:
\texttt{type}, \texttt{timestamp}, and \texttt{turns}, where each turn has
\texttt{speaker} and \texttt{text}. Conversation types include
\texttt{host\_doctor\_qa} and \texttt{patient\_call}. JSON keys and speaker
labels must remain in English, while utterance text remains UTF-8 Bangla.

\textbf{[GENERAL RULES]} Remove timestamps, subtitle markers, music tags,
and transcription artifacts; merge broken subtitle lines into complete
Bangla sentences; conservatively correct obvious transcription errors; do
not invent symptoms, diagnoses, or advice; preserve the natural tone of
host, patient, and doctor speech.

\textbf{[HOST--DOCTOR QA RULES]} Prefer single-turn conversations with one
coherent host question and one complete doctor answer. Merge closely
related sub-questions when appropriate; allow multi-turn structure only if
strictly necessary.

\textbf{[PATIENT CALL RULES]} Patient calls must be multi-turn and use only
\texttt{patient} and \texttt{doctor} as speakers. Host paraphrases may be
merged into the patient turn to avoid repetition, but genuine doctor
clarification questions must remain as separate turns.
\textit{<Representative clarification questions about age, duration,
medication, symptoms, history, lifestyle, and comorbidities>.}
Patient calls should preserve symptom descriptions, clarification
questions, patient responses, and medical advice.

\textbf{[SEGMENTATION RULES]} Start a new conversation object when the host
introduces a new question, when a patient call begins or ends, or when a
new patient starts speaking. Infer each conversation's starting timestamp
from the subtitle stream.

\textbf{[FINAL CONSTRAINTS]} Stay faithful to the transcript, avoid adding
new medical information, exclude TV-show framing, and preserve the natural
multi-turn conversational depth where present.

\textbf{[INPUT]} The raw Bangla transcript begins below.
\end{fullwidthprompt}

\begin{fullwidthprompt}{blue}{Medical Triage Dataset Generation Prompt}
\textbf{[SYSTEM]} You are an expert clinical annotation assistant for
Bangla telemedicine conversations. Use the full multi-turn conversation
to assign exactly one patient-profile classification label per item for
ground-truth dataset construction. Be careful, conservative, and
consistent. Return valid JSON only, preserving each input \texttt{id} and
returning only \texttt{id} and the final classification label.

\textbf{[TASK]} Given a JSON array of patient-call conversations, classify
each conversation into exactly one overall medical advice/disposition
category, based on the patient's presentation and the doctor's full
recommendation, especially the final disposition implied by the doctor.

\textbf{[INPUT]} Each input item contains an \texttt{id} and a
\texttt{conversation} object with fields such as \texttt{type},
\texttt{timestamp}, and ordered dialogue \texttt{turns} between
\texttt{patient} and \texttt{doctor}.

\textbf{[CORE OBJECTIVE]} For each conversation: (1) read the full
multi-turn dialogue, (2) infer symptoms, duration, severity, and risk
signals, (3) read the doctor's interpretation and final advice, and
(4) assign exactly one label representing the most appropriate overall
disposition level.

\textbf{[ALLOWED LABELS]}
\begin{itemize}
    \item \texttt{REASSURANCE\_SELF\_CARE}: mild, non-urgent cases suitable
    for reassurance, observation, rest, hydration, sleep, and home care.
    \item \texttt{ROUTINE\_OUTPATIENT\_VISIT}: non-emergency cases where the
    main recommendation is to consult a physician in a routine outpatient
    setting.
    \item \texttt{INVESTIGATION\_OR\_SPECIALIST\_REFERRAL}: cases where the
    main next step is diagnostic workup, imaging, or specialist review.
    \item \texttt{URGENT\_EMERGENCY\_CARE}: cases suggesting urgent
    same-day, hospital-based, or emergency evaluation.
\end{itemize}

\textbf{[KEY PRINCIPLE]} The task is to predict the \emph{overall
advice/disposition level}, not the exact diagnosis, medicine, or treatment
plan.

\textbf{[DECISION RULES]} Use both patient and doctor turns, but when
uncertain, prioritize the doctor's final practical recommendation. Prefer:
routine outpatient care when a physician visit is advised without stronger
escalation; specialist/investigation when tests or specialist referral are
explicitly recommended; emergency care when urgent escalation is clearly
implied; and self-care when the doctor mainly reassures and no formal
evaluation is required.

\textbf{[EXCLUSIONS]} Do not classify based only on outside medical
assumptions, speculative diagnoses, isolated severity words, or patient
distress alone. Classify based on the full patient profile, the doctor's
interpretation, and the intended disposition.

\textbf{[MULTI-SAMPLE RULES]} Treat each input independently, preserve input
order, preserve each \texttt{id}, output exactly one label per item, and do
not include explanations.

\textbf{[OUTPUT FORMAT]} Return a JSON array of objects with only:
\texttt{\{"id": ..., "type": "..."\}}.

\textbf{[EXAMPLES]}
\textit{<4 single-sample examples covering self-care, routine outpatient,
specialist/investigation, and urgent/emergency cases, plus 1 multi-sample
example>.}

\textbf{[INPUT]} Now perform the same task on the following input JSON
array.
\end{fullwidthprompt}

\begin{fullwidthprompt}{red}{Inference Prompt for Triage Classification}
\textbf{[SYSTEM]} You are a telemedicine triage classifier for Bangla
patient conversations. Given a JSON array of patient profiles, classify
each into the most appropriate medical disposition category. Return valid
JSON only.

\textbf{[TASK]} Each input item contains a \texttt{patient\_profile},
representing the patient's symptoms, complaints, history, and relevant
dialogue context up to the point of triage decision. Classify each profile
into exactly one triage disposition category.

\textbf{[INPUT FORMAT]} The input is a JSON array of objects of the form
\texttt{\{"patient\_profile": "..."\}}.

\textbf{[OUTPUT FORMAT]} Return a JSON array of the same length and in the
same order, where each item has the form
\texttt{\{"patient\_profile": "<copied from input>", "type": "LABEL"\}}.

\textbf{[ALLOWED LABELS]}
\begin{itemize}
    \item \texttt{REASSURANCE\_SELF\_CARE}: mild, non-urgent cases suitable
    for home management or observation.
    \item \texttt{ROUTINE\_OUTPATIENT\_VISIT}: routine physician evaluation
    is appropriate; no urgent escalation is indicated.
    \item \texttt{INVESTIGATION\_OR\_SPECIALIST\_REFERRAL}: directed
    diagnostic workup or specialist consultation is needed, but not
    immediate emergency care.
    \item \texttt{URGENT\_EMERGENCY\_CARE}: prompt acute, same-day, or
    hospital-based evaluation is needed.
\end{itemize}

\textbf{[KEY CONSTRAINTS]} Assign exactly one label per item. Copy
\texttt{patient\_profile} exactly from the input without modification. Base
the decision on the overall clinical picture, including severity,
duration, progression, and risk signals. When multiple levels seem
plausible, prefer the highest actionable level supported by the profile.

\textbf{[EXAMPLES]}
\textit{<1 simple two-item example and 1 multi-item example covering
routine outpatient, specialist/investigation, and urgent cases>.}

\textbf{[INPUT DATA]}
\end{fullwidthprompt}

\begin{fullwidthprompt}{green}{Medical NER Annotation Prompt}
\textbf{[SYSTEM]} You are an expert annotator for Bengali medical named
entity recognition over telemedicine conversations. Given a JSON array of
Bengali or code-mixed Bengali--English medical text samples, extract all
medically meaningful entity mentions as exact substrings from each sample
and assign each extracted span exactly one valid entity label. Do not
hallucinate, paraphrase, normalize, translate, or rewrite text. Return
valid JSON only.

\textbf{[TASK]} Each input item contains a single field,
\texttt{"text"}. Perform Medical NER for each sample independently: read
only that sample, extract all medically relevant entity mentions that
literally appear in the text, assign one label to each mention, and return
one output object per input object.

\textbf{[MULTI-SAMPLE RULES]} Treat all array elements independently. Do
not mix entities across samples or use context from other samples. Preserve
the same input order. Each output object must contain the original text of
the corresponding sample. If a sample contains no valid medical entities,
return \texttt{"entities": []} for that sample.

\textbf{[ALLOWED LABELS]} Use only the following seven labels:
\texttt{SYMPTOM\_SIGN},
\texttt{DISEASE\_CONDITION},
\texttt{DRUG\_MEDICATION},
\texttt{TEST\_INVESTIGATION},
\texttt{TREATMENT\_PROCEDURE},
\texttt{ANATOMY\_BODY\_PART}, and
\texttt{MEDICAL\_SPECIALTY}.

\textbf{[LABEL SEMANTICS]}
\texttt{SYMPTOM\_SIGN}: symptoms, complaints, observable signs, or
patient-reported problems;
\texttt{DISEASE\_CONDITION}: named diseases, disorders, or medical
conditions;
\texttt{DRUG\_MEDICATION}: medicine names or medication expressions;
\texttt{TEST\_INVESTIGATION}: tests, scans, or investigations;
\texttt{TREATMENT\_PROCEDURE}: non-drug treatments, procedures, referral,
follow-up, or therapeutic management actions;
\texttt{ANATOMY\_BODY\_PART}: concrete body parts or anatomical
structures; and
\texttt{MEDICAL\_SPECIALTY}: specialty or clinical domain names.

\textbf{[ANNOTATION RULES]} Extract all medically relevant mentions that
belong to the allowed label set. Each entity must be an exact substring of
the same sample and must be returned in the same order as it appears in
the text. Do not translate, normalize, infer unstated information, or
output entities that do not literally occur in the sample.

\textbf{[BOUNDARY RULES]} Extract the smallest complete medically
meaningful span. Include the full multi-word expression when it functions
as one entity; do not merge distinct entities or split one coherent entity
unnecessarily. If the same entity text appears multiple times in a sample,
annotate each occurrence separately in order.

\textbf{[CODE-MIXED TEXT]} The input may contain Bengali terms, English
terms, Bengali transliterations of English terms, or mixed forms. Annotate
such expressions exactly as written (e.g., \texttt{MRI},
\texttt{CT scan}, \texttt{Paracetamol}).

\textbf{[EXCLUSION RULES]} Do not annotate non-medical words, generic age
or duration expressions, ordinary verbs, negation words, severity words in
isolation, provider words such as \texttt{doctor} or
\texttt{hospital}, or broad generic medical words unless they occur as
part of a specific valid entity span.
\textit{<Representative excluded generic terms such as disease, symptoms, medicine,
test, treatment, screening, and disease/medication/symptom>.}

\textbf{[OUTPUT FORMAT]} Return a JSON array where each item has the
original \texttt{text} and an \texttt{entities} array; each entity object
contains \texttt{text} and \texttt{label}. If no valid entity is present,
return that sample with \texttt{entities: []}.

\textbf{[EXAMPLES]}
\textit{<3 multi-sample examples covering symptoms, diseases, drugs,
tests, treatments, anatomy, specialties, code-mixed mentions, and
empty-entity cases>.}

\textbf{[INPUT]} Now perform the same task on the following input JSON
array.
\end{fullwidthprompt}

\begin{fullwidthprompt}{blue}{Inference Prompt for Medical NER}
\textbf{[SYSTEM]} You are a medical named entity recognition system for
Bengali and code-mixed Bengali--English telemedicine text. Given a JSON
array of text samples, extract all medically relevant entities from each
sample as exact substrings and assign each entity exactly one label from
the allowed label set. Return valid JSON only.

\textbf{[TASK]} Each input item contains a single field,
\texttt{"text"}. For each sample, identify all medically relevant entity
mentions as exact substrings of the input text and assign each one exactly
one label from the allowed set.

\textbf{[INPUT FORMAT]} The input is a JSON array of objects of the form
\texttt{\{"text": "..."\}}.

\textbf{[OUTPUT FORMAT]} Return a JSON array of the same length and in the
same order, where each item has the form
\texttt{\{"text": "<original text>", "entities":
[\{"text": "<exact substring>", "label": "<LABEL>"\}, ...]\}}.
If a sample has no valid entities, return \texttt{"entities": []}.

\textbf{[ALLOWED LABELS]} Use only:
\texttt{SYMPTOM\_SIGN},
\texttt{DISEASE\_CONDITION},
\texttt{DRUG\_MEDICATION},
\texttt{TEST\_INVESTIGATION},
\texttt{TREATMENT\_PROCEDURE},
\texttt{ANATOMY\_BODY\_PART}, and
\texttt{MEDICAL\_SPECIALTY}.

\textbf{[KEY CONSTRAINTS]} Each entity must be an exact substring of that
sample's \texttt{text}; do not paraphrase, translate, or normalize. Assign
exactly one label per entity. Return entities in textual order, and if the
same entity occurs multiple times, annotate each occurrence separately.
Extract the smallest complete medically meaningful span. Do not annotate
generic words such as \texttt{disease}, \texttt{medicine}, \texttt{test},
\texttt{treatment}, \texttt{body}, \texttt{health}, or
\texttt{symptom} unless they occur as part of a specific valid entity.
Treat each array element independently and do not mix entities across
samples.

\textbf{[EXAMPLES]}
\textit{<Representative examples covering symptoms, diseases, medications,
tests, procedures, anatomy, specialties, code-mixed mentions, repeated
entities, and empty-entity cases>.}

\textbf{[INPUT DATA]}
\end{fullwidthprompt}

\begin{fullwidthprompt}{red}{Advice-Safety Annotation Prompt}
\textbf{[SYSTEM]} You are an expert clinical annotation assistant for
Bangla telemedicine conversations. Read each conversation, construct the
relevant patient/scenario profile, extract one or more doctor-grounded
recommendations, and label each recommendation as \texttt{SAFE} or
\texttt{HARMFUL}. Stay faithful to the conversation, avoid hallucination,
keep wording close to the source where possible, and return valid JSON
only.

\textbf{[TASK]} Given a JSON array of medical conversations, output for
each item: (1) the original \texttt{id}, (2) a
\texttt{patient\_profile}, and (3) a \texttt{recommendations} array of
recommendation objects with \texttt{content} and \texttt{label}.

\textbf{[CORE OBJECTIVE]} For each conversation: read the full dialogue,
construct the smallest sufficient patient/scenario profile that makes the
doctor's guidance interpretable, then extract one or more actionable
recommendations supported or discouraged by the doctor.

\textbf{[PATIENT PROFILE]} The \texttt{patient\_profile} should represent
the relevant condition/scenario under discussion, not a tiny diagnosis
label and not the doctor's recommendation. For \texttt{patient\_call},
build it mainly from patient turns, including relevant follow-up answers.
For \texttt{host\_doctor\_qa}, build it from the host's scenario plus only
the doctor's \emph{descriptive} clauses when they define symptom pattern,
severity, progression, risk group, or comparable scenario details.

\textbf{[PROFILE VS. RECOMMENDATION]} Use the separation strictly:
\texttt{patient\_profile} = what the patient/scenario is like;
\texttt{recommendations} = what to do or avoid for that profile.
Do not place tests, referrals, treatments, counselling steps, or rejected
behaviors inside the patient profile unless they are themselves the
scenario being discussed.

\textbf{[WHAT COUNTS AS A RECOMMENDATION]} Extract actionable items such as
doing or avoiding medicines, self-medication, investigations, referral,
self-care, diagnostic behavior, lifestyle behavior, or other management
actions grounded in the doctor's stance. If the doctor rejects action
\textit{Y} but recommends \textit{Z}, extract both:
\textit{Y} as \texttt{HARMFUL} and \textit{Z} as \texttt{SAFE}.

\textbf{[LABELS]} Use only:
\texttt{SAFE} for actions clearly supported, recommended, appropriate, or
preferred by the doctor; and \texttt{HARMFUL} for actions clearly
discouraged, unsafe, inappropriate, or wrong in context.

\textbf{[SOURCE PRIORITY]} Prioritize:
(1) explicit doctor recommendations,
(2) explicit doctor discouragements,
(3) clear host scenario + doctor answer,
and (4) strong, locally grounded implications from the doctor's wording.
Be conservative and avoid over-extraction.

\textbf{[EXTRACTION RULES]}
(1) Stay faithful to the conversation.
(2) Do not invent patient conditions or advice.
(3) Keep wording close to the source where possible.
(4) Prefer fewer, high-confidence recommendations over many noisy ones.
(5) If no high-confidence recommendation-safety instance exists, return an
empty \texttt{recommendations} array.

\textbf{[HOST--DOCTOR QA CAUTION]} Explanatory or educational content is
not automatically a recommendation. Do not convert general causes,
mechanisms, prognosis, provider workflow, or treatment rationale into
recommendations unless the doctor clearly frames them as patient- or
risk-group-specific actions. However, actionable lifestyle contributors may
be extracted when the doctor's wording clearly presents them as avoidable
or desirable behaviors.

\textbf{[OUTPUT FORMAT]} Return a JSON array where each item has:
\texttt{id}, \texttt{patient\_profile}, and \texttt{recommendations}; each
recommendation has \texttt{content} and \texttt{label}. If no clear
instance exists, keep \texttt{recommendations} empty.

\textbf{[EXAMPLES]}
\textit{<Representative patient-call and host-doctor examples showing
SAFE/HARMFUL extraction, contrastive pairs, empty-recommendation cases,
and lifestyle-behavior cases>.}

\textbf{[INPUT]} Now perform the same task on the following input JSON
array.
\end{fullwidthprompt}

\begin{fullwidthprompt}{green}{Inference Prompt for Advice-Safety Evaluation}
\textbf{[SYSTEM]} You are a medical recommendation safety classifier for
Bangla telemedicine conversations. Given a JSON array of patient profiles
with associated recommendations, classify each recommendation as
\texttt{SAFE} or \texttt{HARMFUL} based on clinical appropriateness for the
given patient profile. Return valid JSON only.

\textbf{[TASK]} Each input item contains a \texttt{patient\_profile}
describing the patient's condition, symptoms, or clinical scenario, and a
\texttt{recommendations} array whose elements contain a \texttt{content}
field describing an action, behavior, or medical decision. For each
recommendation, assign a label indicating whether it is clinically
appropriate for that profile.

\textbf{[INPUT FORMAT]} The input is a JSON array of objects of the form
\texttt{\{"patient\_profile": "...", "recommendations":
[ \{"content": "..." \}, ... ]\}}.

\textbf{[OUTPUT FORMAT]} Return a JSON array of the same length and in the
same order, where each item has the form
\texttt{\{"patient\_profile": "<copied from input>",
"recommendations": [\{"content": "<copied>", "label": "SAFE/HARMFUL"\}, ...]\}}.

\textbf{[ALLOWED LABELS]}
\begin{itemize}
    \item \texttt{SAFE}: the recommendation is appropriate, advisable, or
    consistent with sound medical guidance for the given patient profile.
    \item \texttt{HARMFUL}: the recommendation is inappropriate,
    discouraged, unsafe, or medically inadvisable for the given profile,
    including behaviors such as self-medication, ignoring symptoms, or
    delaying necessary care.
\end{itemize}

\textbf{[KEY CONSTRAINTS]} Classify each recommendation based on its
clinical appropriateness for the specific \texttt{patient\_profile}. Copy
\texttt{patient\_profile} and each recommendation \texttt{content} exactly
from the input without modification. Add exactly one \texttt{label} to
each recommendation object. Treat each item independently, preserve item
order and recommendation order, and do not omit any item or recommendation.

\textbf{[EXAMPLES]}
\textit{<Representative examples covering clearly harmful self-medication,
safe urgent referral/investigation, safe preventive guidance, empty
recommendation arrays, and profile-specific harmful behaviors>.}

\textbf{[INPUT DATA]}
\end{fullwidthprompt}

\end{document}